\documentclass[journal]{IEEEtran}

\usepackage{xcolor,soul,framed} 

\colorlet{shadecolor}{yellow}
\usepackage[pdftex]{graphicx}
\graphicspath{{../pdf/}{../jpeg/}}
\DeclareGraphicsExtensions{.pdf,.jpeg,.png}

\usepackage[cmex10]{amsmath}
\usepackage{array}
\usepackage{float}
\usepackage{mdwmath}
\usepackage{mdwtab}
\usepackage{eqparbox}
\usepackage{url}
\usepackage{tikz}
\usepackage{amsmath,amssymb,amsfonts}
\usepackage{algorithmic}
\usepackage{graphicx}
\usepackage{subfigure}
\usetikzlibrary{positioning}
\usepackage[english]{babel}
\usepackage{amsthm}

\theoremstyle{definition}
\newtheorem{definition}{Definition}
\newtheorem{example}{Example}

\theoremstyle{remark}

\begin{document}

\bstctlcite{IEEEexample:BSTcontrol}
    \title{Safety Verification for Evasive Collision Avoidance in Autonomous Vehicles with Enhanced Resolutions}
\author{Aliasghar Arab$^*$,~\IEEEmembership{Senior Member,~IEEE,}
      Milad Khaleghi,~\IEEEmembership{ Member,~IEEE,}
      Alireza Partovi,~\IEEEmembership{ Member,~IEEE,}\\
      Alireza Abbaspour,~\IEEEmembership{ Senior Member,~IEEE,} Chaitanya Shinde,~\IEEEmembership{ Member,~IEEE,} Yashar Mousavi,~\IEEEmembership{ Member,~IEEE,}\\ Vahid Azimi,~\IEEEmembership{Member,~IEEE,} and Ali Karimmoddini,~\IEEEmembership{Senior Member,~IEEE}
\thanks{A. Arab is with the Department of Mechanical and Aerospace Engineering at New York University, Brooklyn, NY, USA and Department of Electrical Engineering at North Carolina A\&T State University, Greensboro, NC, USA, (email: aliasghar.arab@nyu.edu; aarab@ncat.edu), M. Khaleghi is with the Department of Electrical Engineering at North Carolina A\&T State University, Greensboro, NC, USA (email: mkhaleghi@ncat.edu), A. Partovi is an alumni from Department of Electrical and Electronic Engineering of University of Notre Dame (email: apartovi@alumni.nd.edu), A. Abbaspour is Staff System Safety Engineer at Qualcomm (email: alireza.abaspour@gmail.com), C. Shinde is Purdue Alumni and currently work with Torc Robotics, Blacksburg, Virginia, USA \textcolor{black}{(email: cshinde@purdue.edu)}, Y. Mousavi is an Alumni from Department of Applied Science, School of Computing, Engineering and Built Environment, Glasgow Caledonian University, Glasgow, UK and currently works at American Axle \& Manufacturing (email: seyedyashar.mousavi@gcu.ac.uk), V. Azimi is a Tech Lead at Gatik AI, Mountain View, CA, USA (email: vahid.azimi@gatik.ai), and A. Karimoddini is with the Department of Electrical Engineering at North Carolina A\&T State University, Greensboro, NC, USA (email: akarimod@ncat.edu). }
}
\maketitle

\begin{abstract}
This paper presents a comprehensive hazard analysis, risk assessment, and loss evaluation for an Evasive Minimum Risk Maneuvering (EMRM) system designed for autonomous vehicles. The EMRM system is engineered to enhance collision avoidance and mitigate loss severity by drawing inspiration from professional drivers who perform aggressive maneuvers while maintaining stability for effective risk mitigation. Recent advancements in autonomous vehicle technology demonstrate a growing capability for high-performance maneuvers. This paper discusses a comprehensive safety verification process and establishes a clear safety goal to enhance testing validation. The study systematically identifies potential hazards and assesses their risks to overall safety and the protection of vulnerable road users. A novel loss evaluation approach is introduced, focusing on the impact of mitigation maneuvers on loss severity. Additionally, the proposed mitigation integrity level can be used to verify the minimum-risk maneuver feature. This paper applies a verification method to evasive maneuvering, contributing to the development of more reliable active safety features in autonomous driving systems.
\end{abstract}


\IEEEpeerreviewmaketitle
\section{Introduction}
\IEEEPARstart {I}{n} today's fast-paced digital world, driving has evolved from a manual task into a complex information-processing challenge. With the proliferation of digital distractions and a decline in human attention spans, the task of automated and autonomous driving has become increasingly demanding. Evasive maneuvers as a minimum risk maneuver for SAE level 3 and 4 Autonomous Vehicles (AV) offer a promising solution to enhance overall transportation and road safety by compensating for these human limitations. When drivers are overwhelmed by the information they need to process, autonomous systems equipped with advanced evasive maneuvers can step in to mitigate potential risks. These systems utilize cutting-edge technology to rapidly analyze situations and execute maneuvers that are beyond human capabilities, providing a crucial safety net. By integrating these advanced systems, we can significantly reduce the likelihood of accidents, ensuring that vehicles respond to hazards with speed and precision. Identifying autonomous driving limitations offers insights for enhancement opportunities that helps safer adoption to the future proof transportation systems~\cite{Chougule2024Comprehensive}. This paper studies how evasive maneuvers can enhance autonomous vehicle safety by applying a comprehensive safety assessment. 

\noindent Active safety features, such as Advanced Emergency Brake (AEB), are unable to perform Minimum Risk Maneuvers (MRM) in specific hazardous scenarios where braking alone is insufficient to avoid an accident. An MRM is the fallback function by an Automated Driving System (ADS) to reach a stable and stopped state with a minimal risk condition~\cite{iso22736}. While the ISO/SAE Level 3-5 ADS performs the dynamic driving task, an event that prevents the ADS from continuing the dynamic driving task can occur~\cite{iso22736, ISO26262}. An MRM feature aims to improve road safety by focusing on new technologies and strategies to reduce accidents and enhance overall traffic safety~\cite{nayak2024regulatory}. Similar to MRM, evasive maneuvers inspired by professional and stunt car divers can be used to avoid hazardous situations or mitigate harms severity~\cite{arab2024safepatent, arab2021phd}. Integration of an Evasive MRM (EMRM) feature into an Advanced Driving Assistant System (ADAS) or AV will require comprehensive safety Verification and Validation (V\&V)~\cite{grimm2018survey, pang2023survey}.  

\begin{figure}[t!]
	\centering
	\includegraphics[width=3.35in]{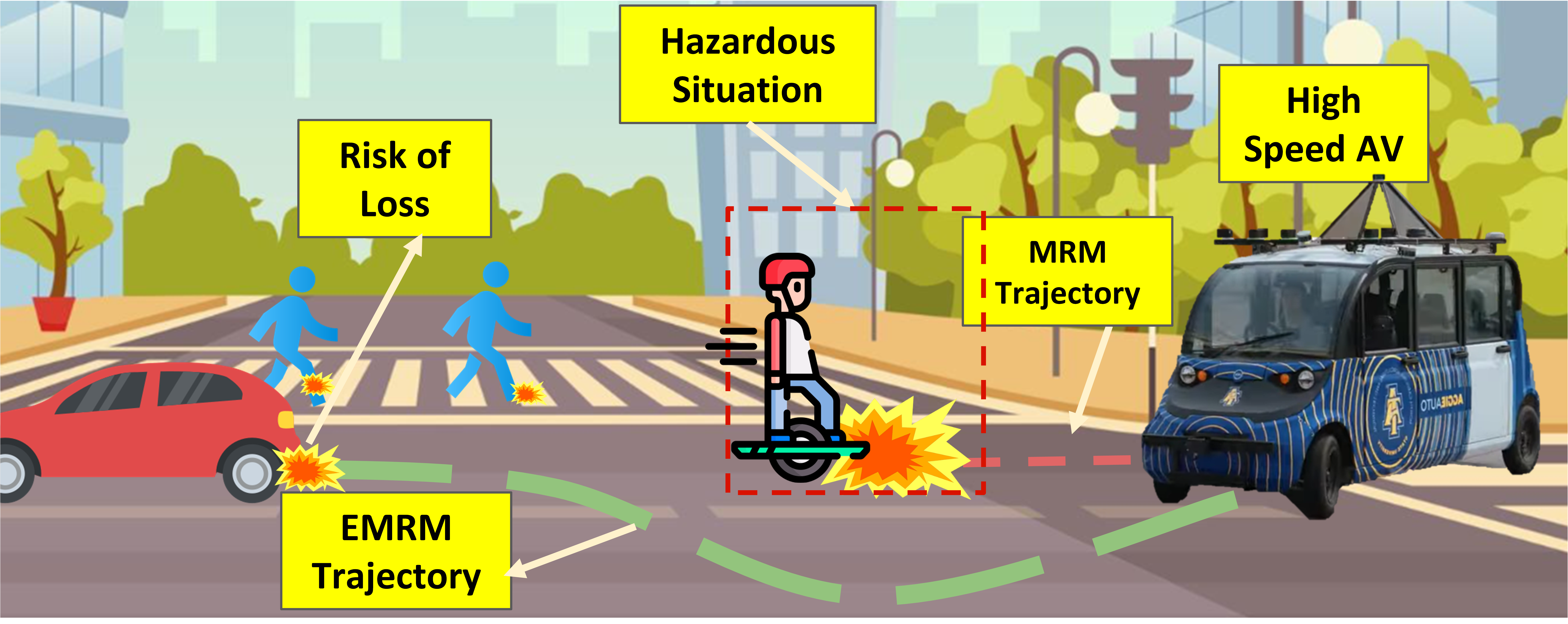}
	\caption{A schematic depicting a hazardous scenario where a basic MRM might fail to avoid or mitigate the loss in an urban operational designed domain. Advanced EMRM can perform aggressive evasive maneuvers to avoid a catastrophic hazard.}
	\label{fig:realscenario}
 \vspace{-0mm}
\end{figure}

\noindent \textcolor{black}{Hazard Analysis and Risk Assessment (HARA) is a crucial first step and popular method for verifying Functional Safety (FuSa) in ADAS and AV within the automotive industry~\cite{ISO26262}. Frameworks provided in standards sometime are very general customization might be needed for specific functionalities~\cite{pang2023survey}. A more effective analyze for functions at edge cases can be done by applying the necessary granularity.} Fig.\ref{fig:realscenario} shows an AV driving fast on an urban domain when a scooter rider suddenly blocks the pre-planned path. It puts the AV in a critical hazardous situation with a high risk of a fatal accident. Efforts to mitigate these risks might not always be successful, potentially leading to other unexpected accidents. Hence, it is crucial to understand the circumstances under which severe hazards are avoidable and the loss level can be reduced. In this scenario, an evasive system can mitigate the situation so that the AV encounters a lower fatality risk, albeit with a higher risk of property or vehicle damage. Consequently, it is crucial that the system assess the situation and choose the path that minimizes harm, whether it be the vehicle, property, life, or health.

\noindent Exposure to hazardous situations and the reduction of crash risk or loss effects involve critical decision-making processes that require high-performance processing and agile responses.~\cite{pawar2021investigating}. When an unexpected emergency occurs, applying acceleration at a rate higher than comfortable driving is inevitable~\cite{ma2020line}. Therefore, performing evasive maneuvers with guaranteed safety can be used for MRMs~\cite{arab2020safety, seo2022safety}. Gleirscher and Kugele~\cite{gleirscher2017hazard}, discussed the process of identifying and mitigating hazards in automated driving systems, though it acknowledges limitations in analyzing evasive maneuvers. To tackle safety assessments for EMRMs, this work proposes a framework that adds the loss severity to HARA for analysis of EMRM features. EMRM is designed to initially perform collision avoidance or risk reduction of severe losses when the accident is inevitable. This paper studies how to verify if EMRMs can improve overall FuSa, thereby there is a potential for development of reliable evasive maneuvering feature for AVs.


\noindent Adding loss evaluation to HARA analysis is needed for the EMRM feature where an accident is inevitable. A classic HARA identifies hazardous events and associated risks of exposure. Describing the circumstances leading to a specific loss severity is needed, and they can be modeled and classified by severity as loss states of the system~\cite{suerken2013model}. In addition, identifying systemic hazards through a structured approach by systematically decoding hazards is possible using System Theoretic Process Analysis (STPA)~\cite{ishimatsu2010modeling}. A unique combination of STPA's strength and Finite State Machines (FSM) is also used for AV safety assessments~\cite{savelev2021finite}. It is noteworthy that none of these methods have addressed understanding the safety goals where multiple loss severity increases the complexity of a hazardous scenario. However, it has been shown that these methods can be customized or combined to address the shortages~\cite{liu2019safe}. 

\noindent \textcolor{black}{In this paper, a systematic approach based on the HARA framework is proposed to assess functions for edge cases, such as MRM features. Our novel approach, adds more granularity to a classic HARA~\cite{ISO26262} to differentiate possible losses in severe and critical situations. This is a necessary step before designing and implementing any EMRM feature. EMRM technique can perform evasive maneuvers for collision avoidance or loss severity reduction~\cite{arab2021instructed, han2023safe}. By incorporating a more granular loss severity, the developed approach evaluates the risks with a higher resolution for a range of minimum and maximum loss severity levels at the hazardous event. Risks of losses with different severity are needed to ensure the system either mitigates the hazard risks or reduces the loss severity~\cite{arab2024safepatent}. In this respect, this method effectively handles different types of loss severity during risk assessment. Similarly, the proposed approach can be used to analyze comprehensively next-generation AV and ADAS active safety features where potential loss of life or severe damages should be considered. The contributions of this paper are summarized as follows:
\begin{itemize}
    \item Adding loss evaluation to extend a classic HARA method in order to enhance the verification granularity resolution around severe conditions.
    \item Introducing a framework with more granularity for higher precision analysis of festoons designed for edge cases. 
    \item Conducting a comprehensive functional safety assessment for evasive minimum risk maneuvering feature, designed to mitigate crash risks or reduce loss severity.
\end{itemize}
}
\noindent The remainder of the paper is structured as follows. In section \ref{sec:Evasive Minimum Risk Maneuvering}, Hierarchical structure and features of EMRM will be explained. Section \ref{sec:HARA Procedure} explains the HARA method used in this paper for verifying the safety of EMRM for collision avoidance and loss mitigation. Additionally, levels of risks and losses will be described, and HARA analysis for EMRM with examples of hazardous scenarios will be provided to verify EMRM safety goals. In Section \ref{sec:Verifying Evasive MRM}, hazards of an evasive minimum risk maneuver feature are analyzed. Finally, Section \ref{sec:conclusion} concludes the paper.

\section*{Acronym Table}
\begin{tabular}{| m{1.4cm} | m{6.3cm}|}
    \hline
    \textbf{Acronym} & \textbf{Definition} \\ \hline
    AV & Autonomous Vehicles \\ \hline
    ADS & Automated Driving System \\ \hline
    ADAS & \textcolor{black}{Advanced Driver Assistance Systems} \\ \hline
    HARA & Hazard Analysis and Risk Assessment~\cite{ISO26262} \\ \hline
    FuSa & Functional Safety \\ \hline
    MRM & \textcolor{black}{Minimum Risk Maneuver}~\cite{UL4600} \\ \hline
    EMRM & \textcolor{black}{Evasive Minimum Risk Maneuver} \\ \hline
    VRU & \textcolor{black}{Vulnerable Road User} \\ \hline
    ODD & \textcolor{black}{Operational Design Domain} \\ \hline
    ASIL & Automotive Safety Integrity Level \\ \hline
    STPA & System Theoretic Process Analysis \\ \hline
    V\&V & \textcolor{black}{Verification and Validation} \\ \hline
    UL & \textcolor{black}{Underwriters Laboratories} \\ \hline
    SAE & Society of Automotive Engineers \\ \hline
    ISO & \textcolor{black}{International Organization for Standardization}  \\ \hline
    V2V & \textcolor{black}{Vehicle-to-Vehicle communication} \\ \hline
    \textcolor{black}{V2I} & \textcolor{black}{Vehicle-to-Infrastructure communication} \\ \hline
\end{tabular}

\vspace{-0mm}
\section{Evasive Minimum Risk Maneuvering Systems}\label{sec:Evasive Minimum Risk Maneuvering}
As AV operations expand into larger operational design domains, their interactions with unpredictable environments will increase, making the need for advanced active safety features and autonomous fail-safe operations inevitable. MRMs are a vital component of the safety framework for autonomous systems, as defined by UL 4600~\cite{UL4600}, focusing on predefined actions to minimize risk and ensure safety during system failures or unexpected conditions. \textcolor{black}{Inspired by the aggressive or stunt driving skills provided by professional drivers are considered a viable strategy for avoiding accidents under emergency conditions, especially in unexpected hazardous situations~\cite{arab2020safety},  controlled evasive maneuvers, known as EMRM, might be the only way to mitigate or even avoid an accident for an AV.} This approach enhances MRM features, ensuring that the AV can execute evasive maneuvers safely and effectively when required to mitigate potential hazards. 

The main goal of EMRM is to design a hazard reaction system that guarantees loss severity mitigation in hazardous situations with the risk of severe losses. Fig.~\ref{fig:Hierarchical flow of EMRM} shows a full list of functions for an EMRM feature ~\cite{arab2024safepatent}. After identifying a hazardous situation, a loss evaluation is performed. Then, based on the results, decision-making and control Planning for EMRM are executed. Finally, the control commands are applied to the system.


\subsection{Identification of Hazard and Real-Time Risk Prediction}
\noindent Identification of hazards involves systematically recognizing potential sources of harm within the operational environment of AVs, such as unexpected obstacles or system malfunctions. Real-time risk prediction leverages advanced sensor data and predictive algorithms to assess and anticipate these hazards, allowing the vehicle to make informed decisions to avoid or mitigate risks proactively. This combination ensures that autonomous systems can dynamically adapt to changing conditions, enhancing overall safety and reliability. The system assumes that the vehicle employs a combination of sensors, perception, and prediction to control its actuators optimally. This includes utilizing Light Detection and Ranging (LIDAR), radar, ultrasonic sensors, and cameras for comprehensive object detection and environmental mapping. 

\noindent Advanced event detection cameras have proven that real-time identification of hazards on public roads is possible~\cite{gehrig2024lowlatency}. The onboard high-performance processing unit is capable of running real-time safety-critical software to process data and control the actuators responsible for steering, braking, and acceleration. While we acknowledge the possibility of sensor and subsystem failures with a certain probability, the safety of intended functionality (SOTIF) for these components is outside the scope of this discussion~\cite{ISO21448}. Additionally, Cybersecurity challenges caused by intrusion for the vehicles equipped with V2V (Vehicle-to-Vehicle) and V2I (Vehicle-to-Infrastructure) communication modules to enhance safety by exchanging safety-related information is out of scope as well and more information can be found here~\cite{Girdhar2023Cybersecurity}. 

\subsection{Risk of Loss Prediction}
Loss severity prediction enhances HARA by allowing EMRM systems to prioritize risks, allocate resources effectively, and design more targeted safety controls. By estimating the potential impact of failures, EMRM systems can focus on mitigating the most severe potential losses and ensure efficient use of safety measures, addressing critical safety issues more effectively. This approach aligns with the STPA framework's~\cite{Stpa2018} emphasis on preventing unacceptable losses, improving overall safety verification processes by focusing efforts where they are most needed.

\subsection{Planning and Control of Evasive Maneuvers}
\begin{figure}[b!]
	\centering
	\includegraphics[width=0.49\textwidth]{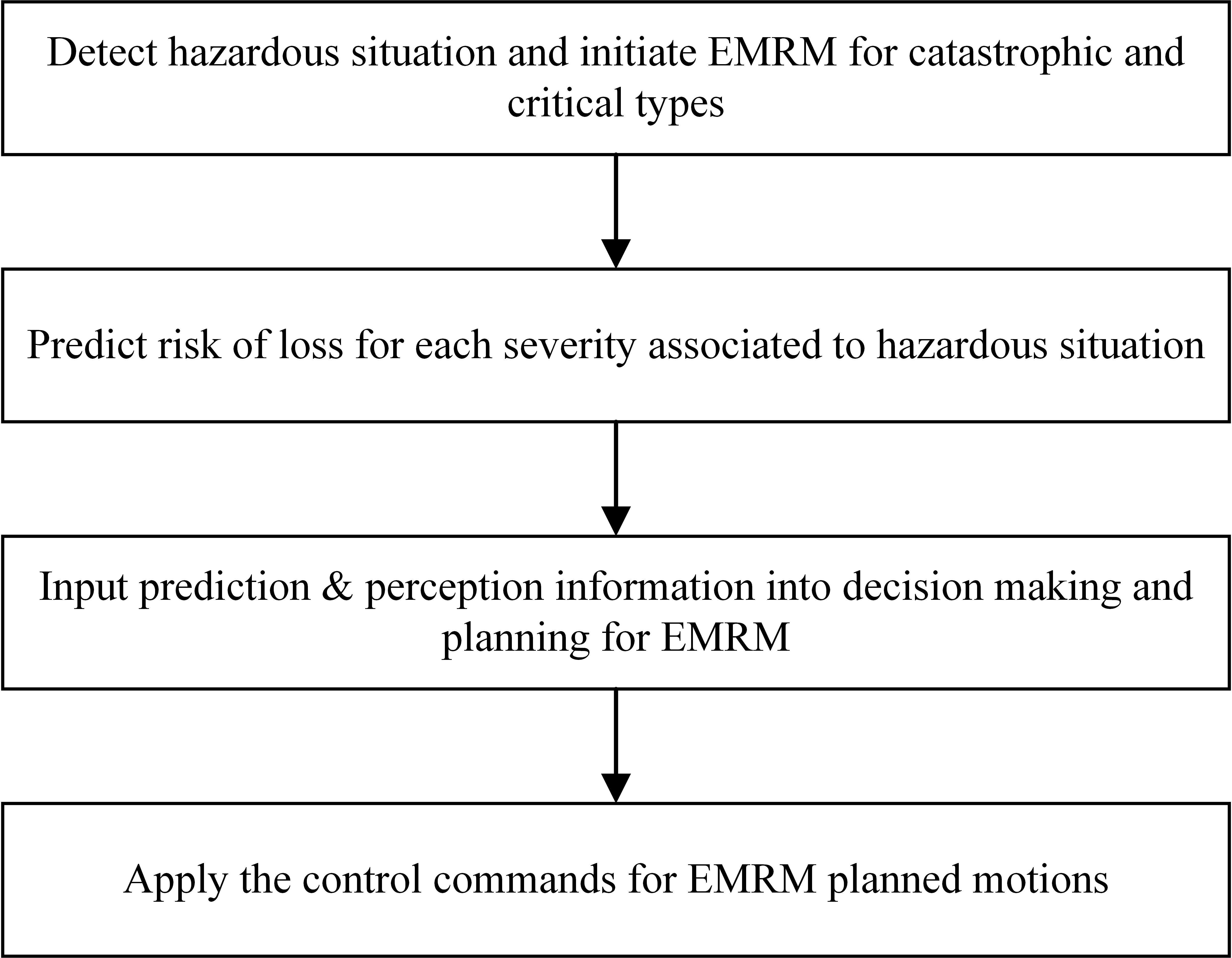}
	\caption{Hierarchical flow of EMRM system.}
	\label{fig:Hierarchical flow of EMRM}
 \vspace{-0mm}
\end{figure}

\noindent Recent research studies demonstrate superior vehicle agility and performance with the potential of EMRM to significantly enhance AV safety in collision and hazard avoidance~\cite{duo2022mpc, li2023planning, zhao2021collision, arab2024motion, nam2019model}. \textcolor{black}{However, while these research articles have shown that evasive maneuvering could potentially improve overall vehicle safety, ensuring the FuSa of EMRMs in the real world becomes paramount.} Hence, the risk mitigation feature must reduce the probability of an accident or alleviate the harm, which is the main objective, not evasive driving. In fact, any MRM feature must guarantee risk and loss mitigation regardless of where, when, and how the feature is used. A high-level framework for an AV with EMRM is shown in Fig.~\ref{fig:Hierarchical flow of EMRM}. A key component for designing a proper EMRM feature is identifying if the vehicle is in a hazardous situation and estimating the risks of loss associated with that situation.


\noindent It is crucial to conduct a thorough safety assessment for EMRM functionalities before testing on real-size vehicles. Previous studies show that by understanding the vehicle dynamic, tire road interaction, and strong perception of the environment, one can perform high-performance maneuvers safely~\cite{arab2021phd, arab2021instructed}. The results demonstrated the potential for being used as advanced safety features for future AVs, especially in emergency hazardous scenarios. Although all the hazardous scenarios where EMRM is needed are not known. HARA analysis allows one to lay out potential risks for early verification of such a feature. Similarly, in practice, all of the potential risks are not identifiable, but it is necessary to list them for verification purposes.

\subsection{System Boundaries}\label{subsubsec:system boundaries}
\noindent The primary objective of EMRM is to create a hazard response system that ensures the reduction of loss severity in potentially dangerous situations with the risk of significant losses. This requires combining analytical methods, data-driven approaches, and machine learning models with probabilistic predictive analysis to ensure that risk mitigation is achievable through evasive maneuvers\cite{lefevre2014survey}. Real-time perception and scene analysis, high-speed planning, and robust control approaches are required to achieve these goals. To effectively assess the loss risk, the comparison for EMRM, MRM, and normal autonomy must be studied both separately and as an integrated system. This independent analysis enables the feasibility study for using EMRM to respond in hazardous situations for loss severity reductions.

\noindent Using these assumptions and definitions, safety assessments for regular MRMs and EMRMs will be possible. However, an EMRM must be executed if and only if a risk of loss exists for any irreversible losses. Once a hazardous state is identified, the MRM module can be activated for avoidance or mitigation. Due to the unpredictability of EMRM performance compared to normal MRMs, EMRM should only be executed in high-risk, severe loss situations where normal MRM fails, and the loss is inevitable for any non-EMRMs. Hence, level of hazards, type of loss and category of the risk and physicsl limits related to hazardous situation are required to identify the effectiveness of an EMRM.

\vspace{-0mm}
\section{HARA Definitions and Procedure}
\label{sec:HARA Procedure}
This section briefly explains the HARA method used in this article for verifying the safety of EMRM for collision avoidance and loss mitigation based on~\cite{ISO26262, UL4600}. Before any new AV feature development, a comprehensive HARA is needed. This ensures that the safety goal is verifiable before the costly integration and deployment phases. Then, the verified safety goal must be validated using rigorous testing and analysis. While a massive number of simulations and real-world tests are necessary to ensure that the EMRM functions correctly under various conditions, initial risk assessments are required to lay out the risks and identify edge-case scenarios for more efficient validation and testing strategies. In this context, a full V\&V report must be available before the feature is deployed in the field for operations. Even though all feasible world states are not known, this is considered a well-accepted initial analysis to identify potential risks. Although, in practice, not all potential risks are identifiable using HARA, it is necessary for informal verification at the early stages of a design process.

\subsection{HARA Taxonomy}\label{subsec:safety verification}
\noindent In this subsection, a modified HARA taxonomy based on ISO 26262~\cite{ISO26262} is outlined to better analyze the MRM feature. This adaptation allows us to specifically evaluate the unique risks and safety measures associated with executing MRM or EMRM in AVs. Here, three key concepts are used to evaluate and prioritize of risks, including Hazard, Exposure, and Controllability.

\subsubsection{Hazard}\label{subsubsec:Hazard}
\noindent A source of harm that can lead to unacceptable loss conditions. Table~\ref{table:list of hazard types} lists the types of hazards, categorized into Catastrophic, Critical, Marginal, and No Impact. Identifying hazard sources and analyzing and avoiding hazardous situations through an algorithmic system is crucial for ensuring the safety and reliability of automated and AVs.
\begin{table}[h!]
\centering
\caption{List of hazard types with descriptions}
\label{table:list of hazard types}
\begin{tabular}{|c|>{\centering\arraybackslash}p{0.15\linewidth}|p{0.64\linewidth}|}
\hline
\textbf{Type} & \textbf{Hazard} & \textbf{Description} \\ \hline
$S_3$ & Catastrophic & Risk of death or serious injury to the public. \\ \hline
$S_2$ & Critical & Risk of major public injuries or serious infrastructure damage. \\ \hline
$S_1$ & Marginal & Risk of minor public injury or property damage. \\ \hline
$S_0$ & No impact & No risk of losses to cause injury to the public or safety-critical systems. \\ \hline
\end{tabular}
\vspace{-0mm}
\end{table}

\subsubsection{Exposure}\label{subsubsec:Exposure}
\noindent The probability of the vehicle being in an operational situation in which a specific hazardous event could occur. Exposure is classified into several categories that represent the probability of occurrence shown in Table~\ref{table:Exposure categories}.
\begin{table}[h!]
\centering
\caption{Exposure classes with probability and descriptions.}
\label{table:Exposure categories}
\begin{tabular}{|c|>{\centering\arraybackslash}p{0.14\linewidth}|p{0.61\linewidth}|}
\hline
\textbf{Class} & \textbf{Exposure} & \textbf{Description}\\ \hline
E$_4$ & High & Highly likely to occur; expected to happen frequently. \\ \hline
E$_3$ & Medium & More likely to occur than not; frequent occurrence. \\ \hline
E$_2$ & Low & Could occur; moderate likelihood of happening. \\ \hline
E$_1$ & Very Low & Low probability of occurrence; not expected but possible. \\ \hline
E$_0$ & Incredible & Incredibly low chance of occurring; exceptional or rare events. \\ \hline
\end{tabular}
\end{table}

\subsubsection{Controllability}\label{subsubsec:Exposure}
The capability of MRM to either avoid a specific hazard event or mitigate the consequences once the event has occurred. As shown in Table~\ref{table:Controllability categories}, the controllability of the MRM system in an AV is classified in four levels: Controllable, Simple, Moderate, and Difficult. Each level reflects the MRM system's capability to manage the risk of harm associated with a hazardous event.
\begin{table}[h!]
\centering
\caption{Levels of Controllability for the MRM system.}
\label{table:Controllability categories}
\begin{tabular}{|c|>{\centering\arraybackslash}p{0.19\linewidth}|p{0.58\linewidth}|}
\hline
\textbf{Cat.} & \textbf{Controllability} & \textbf{Description}\\ \hline
\textcolor{black}{C$_3$} & \textcolor{black}{Difficult} & \textcolor{black}{The hazardous event is challenging to control or uncontrollable with a high risk of harm.} \\ \hline
\textcolor{black}{C$_2$} & \textcolor{black}{Moderate} & \textcolor{black}{The hazardous event can generally be controlled with a medium risk of harm.} \\ \hline
\textcolor{black}{C$_1$} & \textcolor{black}{Simple} & \textcolor{black}{The hazardous event can be easily and reliably controlled with no significant risk of harm.} \\ \hline
\textcolor{black}{C$_{0}$} & \textcolor{black}{Controllable} & \textcolor{black}{The hazardous situation can be avoided and controlled by gentle maneuvers.}\\ \hline
\end{tabular}
\end{table}


\vspace{-0mm}
\subsection{System Definitions}
\label{sec:system definitions}
The proposed method enhances the initial step of HARA as outlined in ISO 26262~\cite{ISO26262} by emphasizing a comprehensive yet streamlined definition of the item being evaluated. The definition focuses on the EMRM autonomy functions, aiming to make the hazard analysis both actionable and focused without being burdened by unnecessary complexities related to excessive functions or sub-functions. In the item definition, we assume perception, prediction, control, planning, and vehicle integrity are highly reliable for analyzing functionality.

\subsubsection{Function Extraction}\label{subsec:function extraction}
Once the item is clearly defined, the functions are identified with contributions from experts in stunt driving. A detailed list of scenarios is then created, evaluating their relevance to the item's requirements and ensuring they reflect the latest developments to mitigate oversight risks. These scenarios are meticulously categorized to cover a spectrum of operational conditions.

\subsubsection{Malfunction Derivation}
\label{subsubsec:malfunction derivation}
After extracting the functions, the next step involves identifying potential malfunctions. This stage requires assessing the relevance of different types of faults to each function and their impact on the system. Crucially, not every fault type will apply to all functions, necessitating a selective approach to evaluation.

\subsection{Hazardous Scenarios}\label{subsec:Hazard Descriptions}
\noindent After identifying malfunctions, the process quickly moves to pinpoint associated hazards, emphasizing their potential to cause harm from an operational standpoint, considering worst-case scenarios. The focus is on behaviors resulting from malfunctions that are perceptible to EMRM features, including \textcolor{black}{Vulnerable Road Users} (VRUs) behavior predictions. Each malfunction is examined to uncover all possible hazardous scenarios for urban operational design domains with different road surface conditions. The aim is to clarify how malfunctions in EMRM translate into observable, potentially more hazardous vehicle behaviors. This analysis will be integrated into the automotive architecture representation, detailing operational situations, hazards, hazardous events, and safety goals, facilitating further item refinement in the development stages.

\subsection{Risk Assessment}\label{subsubsec:remaining risks} 
\noindent According to ISO 26262~\cite{ISO26262}, risk is defined as a combination of the probability of occurrence of harm and the severity of that harm. The assessment helps in determining the necessary Automotive Safety Integrity Level (ASIL) required to ensure that appropriate safety measures are implemented to mitigate the identified risks.  The last step is assessing the feasibility and effectiveness of EMRM as a potential risk mitigation strategy through ASIL. This evaluation includes analyzing the EMRM features' ability to prevent or mitigate the identified risks, ensuring that the implemented strategies can significantly reduce the likelihood and impact of loss in hazardous states. This step determines the adequacy of safety measures and ensures that the system's risk levels are acceptable after performing EMRM. Using qualitative measures for \textit{Severity}, \textit{Exposure}, and \textit{Controllability} of any hazardous scenarios, one can determine the ASIL level and construct a testable safety goal.

\subsection{Definition of Safety Goals} \label{Definition of Safety Goals}
In this research, the safety goal is to clearly defined using human experts in the field as a measurable statement that aims to mitigate risks to an acceptable level \cite{ISO26262}. However, Generative Artificial Intelligence (AI) tools or AI co-pilots are being integrated for standard frameworks~\cite{abbaspour2024enhancing} which can be used for future studies. For example, in an AV's emergency pull-over feature, a safety goal might be to prevent unintended acceleration. This involves identifying scenarios where unintended acceleration could occur, assessing the risk and severity of these scenarios, and implementing control measures such as redundant braking systems and continuous monitoring of acceleration inputs to mitigate the risk.


\section{Applying HARA to EMRM}\label{sec:Verifying Evasive MRM}
In this section, the HARA procedure explained in Sec.~\ref{sec:HARA Procedure} is applied to an EMRM feature explained in Sec.~\ref{sec:Evasive Minimum Risk Maneuvering}. \textcolor{black}{This section will conduct a full HARA analysis for EMRM starting from the system definition.} General HARA procedure and taxonomy is explained in Section~\ref{sec:HARA Procedure} while the definitions are slightly adjusted for EMRM analysis.

\subsection{Functions and Malfunctions of EMRM system}
As Table ~\ref{table:EMRM System Definitions for HARA} shows, four functions are listed for EMRM system as the item definitions and functions are based on the EMRM features explained in Sec.~\ref{sec:Evasive Minimum Risk Maneuvering}. A specified Function ID (FID) is devoted to each function item. \noindent  One can extract malfunctions rooted in any of the functions where they can be identified by their Malfunction ID (MFID). EMRM "Malfunctions Extraction" involves identifying and analyzing potential failures within the EMRM system.  The potential EMRM malfunctions are extracted from system definition and functionality. Specified FID and MFID helps tracing a hazard or risk to the item definitions and product features. This process aims to pinpoint specific malfunction scenarios to enhance EMRM design for reliability and effectiveness in mitigating severe risks during AV operations. 

\begin{table*}[h!]
\centering
\caption{EMRM System Definitions, functions and malfunctions using HARA method.}
\label{table:EMRM System Definitions for HARA}
\begin{tabular}{|c|>{\centering\arraybackslash}p{0.4\linewidth}|c|p{0.17\linewidth}|c|p{0.17\linewidth}|}
\hline
\textbf{Item} & \textbf{Definition} & \textbf{FID} & \textbf{Function} & \textbf{MFID} & \textbf{Malfunction}\\ \hline
I & EMRM System must identify hazardous situations according to types in Table~\ref{table:list of hazard types} & F1 & Real-time hazard detection & \begin{tabular}[c]{@{}l@{}}MF1\\ MF2\\ MF3\end{tabular} & \begin{tabular}[c]{@{}l@{}}No detection\\ Wrong detection\\ Late detection\end{tabular} \\ \hline
II & EMRM System should predict risks for all possible losses & F2 & Risk of loss predictions & \begin{tabular}[c]{@{}l@{}}MF1\\ MF2\\ MF3\end{tabular} & \begin{tabular}[c]{@{}l@{}}No prediction \\ Less severe prediction\\ More severe prediction\end{tabular} \\ \hline
III & EMRM System can plan and decide to perform safety-guaranteed aggressive maneuvers for collision avoidance \cite{arab2020safety}. & F3 & collision avoidance & \begin{tabular}[c]{@{}l@{}}MF1\\ MF2\\ MF3\\ MF4\\ MF5\end{tabular} & \begin{tabular}[c]{@{}l@{}} No evasive maneuver\\ Wrong evasive maneuver\\ Less aggressive maneuver\\ Late evasive maneuver\\ More aggressive maneuver\end{tabular}\\ \hline
IV & EMRM System must alleviate loss severity for inevitable collisions & F4 & Accident impact alleviation & MF1 & Aggravation \\ \hline
\end{tabular}
\end{table*}

\subsection{Listing Hazardous Scenarios}\label{subsec:Hazardous Scenario example}
\noindent Based on malfunctions, hazardous scenarios are identified. Malfunctions such as sensor errors, algorithm failure, or functional issues are analyzed. \textcolor{black}{Each hazardous scenario is identifiable by its Hazard ID (HID)} and one can trace it down to the malfunctions and functions using the Root codes shown in Table~\ref{table:List of hazardous scenarios}. Urban Operational Design Domains (ODDs) are considered to understand the varied risks each scenario might pose in complex situations. This systematic approach ensures a comprehensive set of hazardous scenarios for effective risk assessment. In the hazardous scenario step, \textit{severity} measures the loss severity level, \textit{Exposure} assesses how frequently a particular hazard could occur. \textit{Controllability} evaluates the EMRM's ability to avoid or mitigate the hazard which is explained in an example.

\begin{example}
In Table~\ref{table:list of hazard types}, the hazardous scenario, $H_6$, is the case where EMRM executed an action that results in aggressive motions due to avoiding marginal hazard mistakenly predicted as a severe loss involved.
\begin{itemize}
\item \textbf{Exposure}: The AV is rarely exposed to this malfunction and potential hazards, occurring under specific and infrequent conditions (E$_0$).
\item \textbf{Controllability}: In the event of an erroneous execution of EMRM, the embedded obstacle avoidance system will effectively prevent accidents but due to aggressive maneuvering, the controllability of the vehicle is low (C$_2$).
\end{itemize}
\noindent Using the above-mentioned example as a reference and the malfunctions and system definitions, one can proceed to list all the hazard scenarios as shown in Table~\ref{table:List of hazardous scenarios}.
\end{example}
\subsection{ASIL Results}\label{subsec:ASIL Levels} 
\noindent Three factors determine the ASIL requirement for a particular system: Severity, Exposure, and Controllability. Severity assesses the potential safety consequences on drivers, passengers, or nearby pedestrians and vehicles if the system fails shown in Table~\ref{table:list of hazard types}. Exposure measures the likelihood of an operational situation that could be hazardous in conjunction with the failure mode under analysis listed in Table~\ref{table:Exposure categories}. Controllability evaluates the ability to avoid harm through timely reactions from those involved in the operational situation explained in Table~\ref{table:Controllability categories}. Using the ISO 26262 ASIL level lookup table, some of the scenarios in Table~\ref{table:List of hazardous scenarios} fall into the \textit{Quality Management} (QM) meaning it does not require an ASIL level but should still be managed through proper quality control processes or additional safety assessments to complete HARA.

\subsection{Safety Goal Extraction}\label{subsec:Safety Goal Extraction}
Safety goal extraction in ISO 26262~\cite{ISO26262} involves identifying the top-level safety objectives derived from a thorough hazard analysis and risk assessment. It is needed because these safety goals form the foundation for designing and implementing safety measures that mitigate or eliminate potential hazards, ensuring the system meets the necessary safety integrity levels and complies with functional safety standards. We have extracted the following items as our safety goals for an EMRM feature in AVs.

\begin{itemize}
    \item EMRM systems \textbf{should} assess the situation and make the decision in a timely manner.
    \item EMRM systems \textbf{can} respond to critical and catastrophic hazardous situations.
    \item EMRM systems \textbf{should} reduce the loss severity in such scenarios.
    \item EMRM systems \textbf{might} avoid a hazard.
\end{itemize}
 \noindent Out of 12 hazardous scenarios, only 3 are not critical hazards, H2, H4, and H6 where the aggressive maneuver might be performed mistakenly. However, exposure to scenarios is low and the probability of severe loss is still low. Hence, this will need further analysis on the rest of the hazardous scenarios, even the ASIL is A or QM, these are because the exposures are very low but an MRM is designed for the edge cases and rare events. Safety goals must be established and are necessary to achieve the desired "safe design" as the outputs of the HARA. Since the exposure to these hazardous scenarios is very low, we will not be able to use the ASIL levels to finalize our safety goal using a standard HARA framework for EMRM analysis. MRM systems safety goal is to bring the vehicle to a minimum risk condition. At the same time, aggressive driving is considered a high-risk maneuver. The decision-making for the trade-off between using high-risk maneuvers to reduce the risk of severe injuries can be formulated as constrained optimal control problem with the following objectives, defined as safety goals.

\begin{table*}[h!]
\centering
\caption{List of hazardous scenarios and \textit{Severity}, \textit{Exposure}, and \textit{Controllability} analysis for EMRM system malfunctions in level 4 AVs. The vehicle is analyzed for urban ODDs and the risk assessments are based on ISO 26262 ASIL table.}
\label{table:List of hazardous scenarios}
\begin{tabular}{|c|c|>{\centering\arraybackslash}p{0.61\linewidth}|c|c|c|c|}
\hline
\textbf{HID} & \textbf{Root} & \textbf{Vehicle Level Hazardous Scenario} & \textbf{Sev.} & \textbf{Exp.} & \textbf{Con.} & \textbf{ASIL} \\ \hline
H1 & F1-MF1 &  Evasive MRM not executed, causing a collision due to the failure to detect vulnerable road users(VRUs)~\cite{Calvillo2023dmv}. & S$_3$ & E$_0$ & C$_3$ & QM \\ \hline
H2 & F1-MF2 & EMRM executed, causing unnecessary aggressive maneuvers due to mistakenly detecting a VRU on the road that does not exist. & S$_1$ & E$_0$ & C$_0$ & QM \\ \hline
H3 & F1-MF3 & EMRM not executed, causing a collision with obstacles due to late detection of VRUs on the road due to appearing from blind spots. & S$_3$ & E$_2$ & C$_2$ & A\\ \hline
H4 & F2-MF1 & EMRM not executed, causing an accident with ego vehicles due to inability of hazard and loss risk prediction~\cite{Tangermann2024Yahoo}. & S$_2$ & E$_1$ & C$_1$  & QM \\ \hline
H5 & F2-MF2 & EMRM not executed causing an accident with VRUs due to predicting of a hazard as non severe loss. & S$_3$ & E$_0$ & C$_3$ & QM\\ \hline
H6 & F2-MF3 & EMRM executed, causing aggressive motions due to avoiding marginal hazard mistakenly predicted as a severe loss involved. & S$_1$ & E$_1$ & C$_3$ & QM\\ \hline
H7 & F3-MF1 & EMRM executed, unable to avoid a collision with VRUs due to lack of agility performance. & S$_3$ & E$_1$ & C$_2$ & QM \\ \hline
H8 & F3-MF2 & EMRM executed, causing an accident due to inability to plan a safe evasive maneuver in real-time. & S$_3$ & E$_0$ & C$_3$ & A\\ \hline
H9 & F3-MF3 & EMRM executed, causing an accident due to slow maneuvering performance. & S$_3$ & E$_1$ & C$_2$ & QM\\ \hline
H10 & F3-MF4 & EMRM executed, causing an accident due to computational delay in performing the maneuver. & S$_3$ & E$_2$ & C$_3$ & B\\ \hline
H11 & F3-MF5 & EMRM executed, performing extremely evasive maneuvers and loses the vehicle stability~\cite{yi2011stability}. & S$_3$ & E$_1$ & C$_3$ & A\\ \hline
H12 & F4-MF1 & EMRM executed properly, increasing the loss severity or unable to avoid the hazard due to physical limitation of the system or situation. & S$_3$ & E$_1$ & C$_3$ & A \\ \hline
\end{tabular}
\end{table*}

\section{Loss Evaluations}\label{sec:Loss Severity Reduction Analysis}
\noindent This section introduces loss severity assessments as a critical augmentation to HARA for safety verification of the EMRM systems. \textcolor{black}{Unlike traditional HARA, this approach defines new evaluation variables to assess risks across varying loss severities, especially in edge-case minimum risk maneuvers~\cite{arab2024safepatent}. By addressing a range of impacts, it ensures that risks are mitigated or potential losses minimized, offering a more detailed and comprehensive risk analysis.} The safety goals discussed in Sec.~\ref{subsec:Safety Goal Extraction} mean an EMRM system should mitigate potential losses in critical and catastrophic hazardous scenarios. Given the low probability of exposure to such high-risk situations in Table~\ref{table:List of hazardous scenarios}, the ASIL level can not suffice to find the remaining risks. This section defines a new assessment strategy to verify the EMRM systems are designed to reduce the loss severity. By incorporating new definitions maneuvering confidence, avoidance complexity, and mitigation effectiveness. Evaluating these in each scenario is necessary to assess EMRM system functionality. Minimizing potential damages in each scenario is discussed for various types of vehicle drivers, from humans to Level 4 drivers equipped with EMRM system. Comparing the expected performance using the defined parameters allows one to verify EMRM will enhance overall vehicle and road safety.

\subsection{Loss Definition}
Loss is the actual undesirable outcome or damage where hazard is a precursor condition or state that can lead to a loss. System-Theoretic Process Analysis (STPA) method uses an ODD-based granular list of loss levels for evaluations~\cite{Stpa2018}. Inspired by STPA method, the extracted loss level table is used to evaluate the loss mitigation of the EMRM system outlined in Section~\ref{sec:Evasive Minimum Risk Maneuvering}. Since, EMRM systems are only for critical and catastrophic hazardous situations, less severe losses are not discussed in this section. The following definitions are needed to evaluate the loss mitigation of EMRM systems. 

\subsubsection{Harm or loss severity levels}
Classifying harm or losses based on severity is necessary to analyze the likelihood of EMRM in reducing the overall impact of an accident and mitigating harm severity in an uncontrollable catastrophic hazardous situation.

\begin{definition}[Loss Severity] \label{def:Loss Severity}
Loss or harm severity is defined based on the potential consequences of a hazardous situation in terms of physical injury or damage to the health of people. The severity levels categorize the extent of harm that could result from the event.
\end{definition}

\noindent A List of losses with different severity levels typically includes potential negative outcomes classified by their impact on safety, operations, and mission objectives. These losses range from minor customer dissatisfaction, such as slight discomfort due to high acceleration motions, to severe consequences, including significant injury or fatality to personnel described in Table~\ref{table:Loss Levels}. By categorizing these losses according to severity and integrating them with the potential for mitigation using EMRM, alongside assessing the likelihood of occurrence for each loss, one can effectively evaluate the overall impact of EMRM on the safety of AV operations.

\begin{definition}[Loss State]\label{def:Loss event}
A loss state is any unaccepted state that is the result of an accident event (\textit{Accident}) in a hazardous situation.
\end{definition}
\noindent Loss states are compassing any situation that is not desirable. Hence, classifying losses based on their severity is necessary to understand if an EMRM is reducing the risk for severe losses in a high-risk hazardous situation.

\begin{table}[hb!]
\centering
\caption{Granular Levels of Possible Harms for AVs at Catastrophic or Critical Hazardous Scenarios.}
\label{table:Loss Levels}
\begin{tabular}{|c|>{\centering\arraybackslash}p{0.22\linewidth}|p{0.52\linewidth}|}
\hline
\textbf{Level} & \textbf{Loss} & \textbf{Description} \\ \hline
L$_0$ & Damage to Objects Outside of the Vehicle & Property damage or objects in the environment, resulting in repair costs and potential liabilities. \\ \hline
L$_1$ & Damage with Major Economic Loss & Significant damage to property or AV itself, leading to substantial repair or replacement costs and downtime. \\ \hline
L$_2$ & Serious Damage to Infrastructure & Serious damage to infrastructure causing long-term effects. \\ \hline
L$_3$ & Minor injuries to People & without causing long-term detrimental effects but leading to legal, ethical, and financial consequences. \\ \hline
L$_4$ & Severe Injury to People & Harm to individuals caused by the AV system, leading to severe legal, ethical, and financial consequences. \\ \hline
L$_5$ & Indirect Cause of Loss of Life & Fatality indirectly caused by the AV system, improper AV behaviour leading other vehicles into accident. \\ \hline
L$_6$ & Causing Loss of Life & Fatality directly caused by the AV system, leading to greater severe legal, ethical, and socio-economical consequences. \\ \hline
\end{tabular}
\vspace{-0mm}
\end{table}

\subsection{Loss Evaluation Parameters}\label{subsec:Loss Evaluation Parameters}
\noindent The safety goal for EMRMs in AV is to clearly define a measurable objective that aims to mitigate high-severity loss risks to a lower level of severity losses. For instance, in the context of AV scooter rider avoidance, the safety goal is to prevent unintended collisions with the scooter and not cause higher or similar risks for losses with the same or higher severity. This involves identifying scenarios where other losses could occur, assessing the risk and severity of these scenarios, and implementing control measures such as redundant systems and continuous monitoring of perception and actuation inputs to mitigate the risk effectively.

\subsubsection{Maneuverability}
Maneuverability refers to the capability of a vehicle and planning and control systems to execute safe aggressive maneuvers effectively. This incorporates the physical and functional aspects and the level of confidence in detecting and performing such actions to avoid hazards or mitigate the loss in hazardous situations. Compared with human drivers as a baseline, the results indicate that younger drivers exhibit greater maneuverability in response to hazards due to their quicker detection and decision-making times. Specifically, older drivers (aged 55 to 69) required 403 milliseconds to detect hazards in videos and 605 milliseconds to decide on an avoidance maneuver. In contrast, younger drivers (aged 20 to 25) needed only 220 milliseconds to detect hazards and 388 milliseconds to choose an avoidance action. \textcolor{black}{This suggests that EMRMs are designed to outperform younger drivers, whose performance is considered the baseline for effective and swift hazard avoidance~\cite{matheson2019study}.} Hence, the maneuverability capability is classified in three different levels, explained in Table~\ref{table:Maneuverability}.

\begin{table}[ht!]
\centering
\caption{Maneuverability capability level.}
\label{table:Maneuverability}
\begin{tabular}{|c|>{\centering\arraybackslash}p{0.10\linewidth}|p{0.68\linewidth}|}
\hline
\textbf{Lev.} & \textbf{Skill} & \textbf{Description}\\ \hline
M$_1$ & Basic & The AV can only perform basic maneuvers, similar to older drivers.\\ \hline
M$_2$ & Average & The AV can perform basic maneuvers and also an MRM system is capable of handling moderately hazardous situations, similar to average drivers.\\ \hline
M$_3$ & Pro & The AV can perform basic maneuvers and EMRM system is capable of avoiding critical hazards or mitigating the loss, inspired by professional race-car or stunt drivers~\cite{arab2021phd}.\\ \hline
\end{tabular}
\end{table}

\subsubsection{Avoidability}
Avoidability is the likelihood that a hazardous situation can be effectively avoided or mitigated once it occurs. The Avoidability probability are ranged from Avoidable (A$_1$), Challenging (A$_2$) and Inevitable (A$_3$). These factors can be a parameter of road condition, VRU factors in the scene and the ego vehicle states relative to the VRUs.

\subsubsection{Mitigatability}
Mitigatability is the likelihood that a mitigation maneuver if successful will effectively reduce the loss level and impact of an accident as explained in Table~\ref{table:Loss Levels}. Table~\ref{table:Mitigatability} shows categories to evaluate the loss where mitigablity is calculated based on the probability of reducing a loss is $Prob(M)$ and levels of loss reduction is $LLR$, one can calculate mitigablity as

\begin{equation}\label{eq:emrmrisk}
    \text{Mitigablity} = M \times Prob(M) \times LLR,
\end{equation}

\begin{multline} \label{eq:mitigatability}
    \text{EMRM Risk} = \\ \frac{\text{Baseline Loss} - \text{Mitigablity}}{\text{Baseline Loss}} \times 100.
\end{multline}
\noindent where the baseline loss of the scenario is an AV without the EMRM intervention or an average driver. This can be calculated as the product of Severity (S), Probability (P), Exposure (E), and Loss Severity (L) at that scenario as
\begin{equation}\label{eq:baseline}
    \text{Baseline Loss} = (S \times P \times E \times L).
\end{equation}

\begin{figure*}[h!]
		\hspace{-3mm}
		\centering
		\subfigure[]{
	\label{fig:waymoStreet}
    \includegraphics[width=3.5in,height=2.2in]{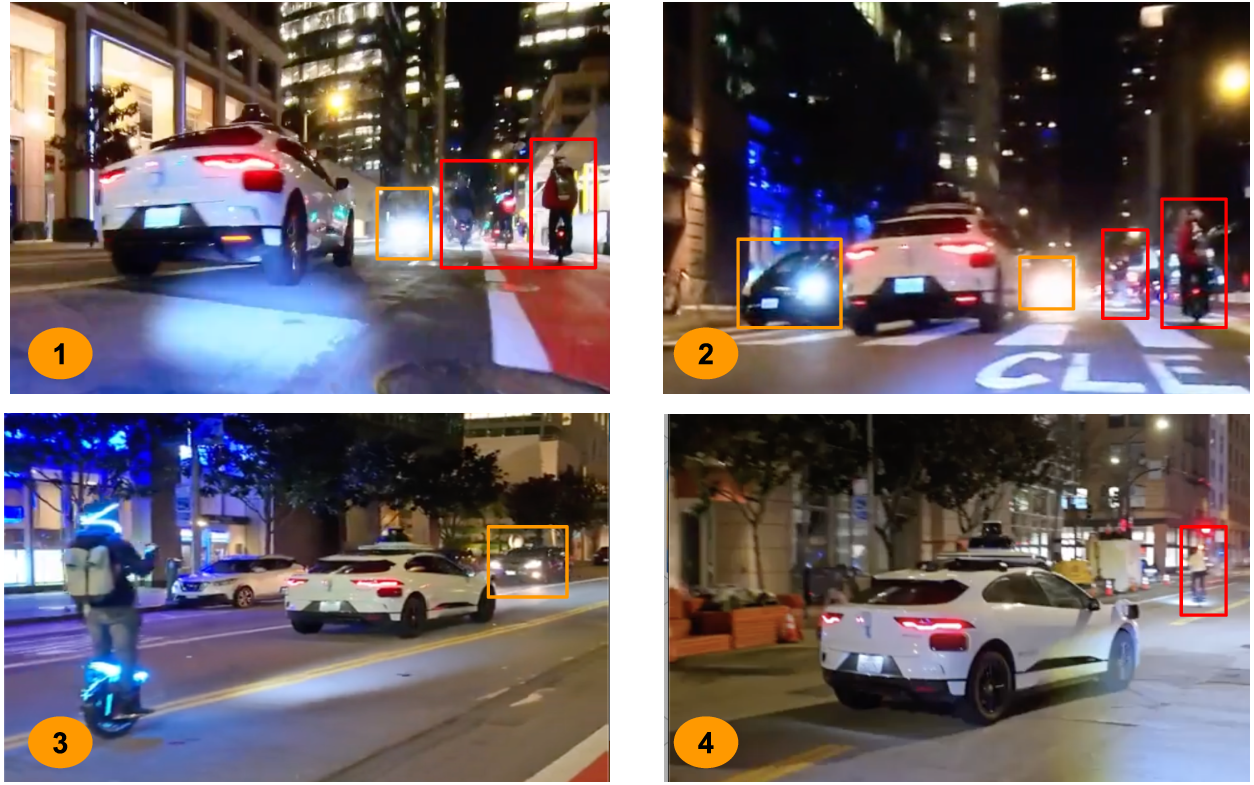}}
		\subfigure[]{
	\label{fig:WaymoSchematic}
    \includegraphics[width=3.5in,height=2.2in]{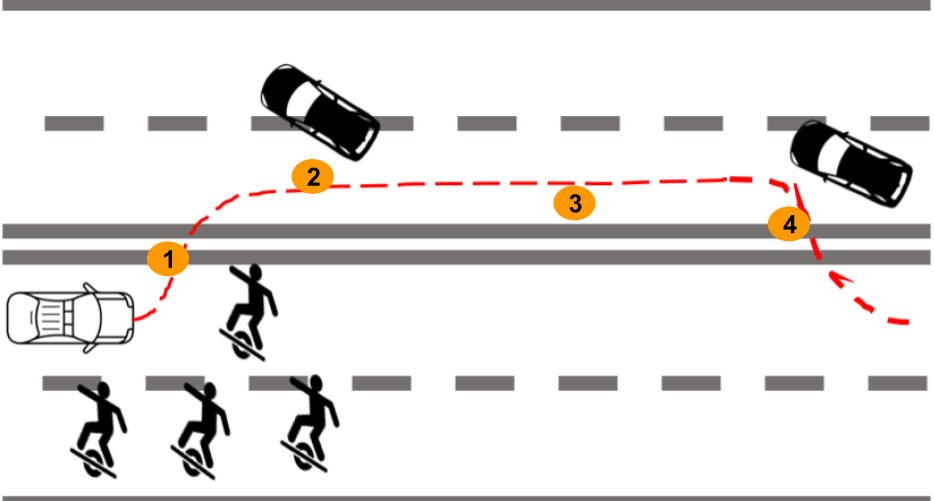}}
	\caption{\textcolor{black}{AV crosses double solid lanes to mitigate a potential collision with a vulnerable road users (scooter riders) to mitigate a potential loss. a) Images captured from videos posted on social media showing an AV is crossing double solid lanes and not returning. b) A schematic of the AVs evasive maneuver for avoiding potential collisions at the same scenario.}}
		\label{fig:waymo}
\end{figure*}

\begin{example}[]\label{example:mitigable}\textcolor{black}{
In a hazardous situation where an AV encounters scooter riders suddenly veering into its path, the AV must quickly decide whether to stay in its lane and risk a collision with the scooter or swerve and cross double solid lines to avoid riders, potentially increasing the chance of hitting another vehicle in the adjacent lane shown in Fig.~\ref{fig:waymo}. The AV chooses to swerve, mitigating the overall loss by prioritizing the scooter rider's safety, who is more vulnerable, even though this maneuver introduces a lesser risk of a side-swipe with another vehicle. This decision reflects the AV's ability to minimize harm in an otherwise unavoidable scenario, demonstrating that the situation was mitigable. Numbers on the images show the steps the vehicle took as an evasive maneuvers. 1) AV crosses double solid lanes. 2) AV blocks a coming vehicle and force them to change their path 3) AV continues driving on the opposite side for 39 seconds. 4) AV drove back to correct lane.}
\end{example}

\begin{table}[h!]
\centering
\caption{The mitigablity category table.}
\label{table:Mitigatability}
\begin{tabular}{|c|>{\centering\arraybackslash}p{0.10\linewidth}|p{0.68\linewidth}|}
\hline
\textbf{Cat.} & \textbf{Impact} & \textbf{Description}\\ \hline
$\mathcal{M}_1$ & Low & Low impact mitigation with a low chance of reducing small levels of loss.\\ \hline
$\mathcal{M}_2$ & Medium & Medium impact mitigation with a medium chance of reducing a few levels of loss.\\ \hline
$\mathcal{M}_3$ & High & High impact mitigation with a good chance of reducing a multiple levels of loss.\\ \hline
\end{tabular}
\end{table}

\begin{table}[b!]
	\centering
 \caption{Loss evaluations for ASIL QM hazardous scenarios in Table~\ref{table:List of hazardous scenarios}.}
	\includegraphics[width=0.49\textwidth]{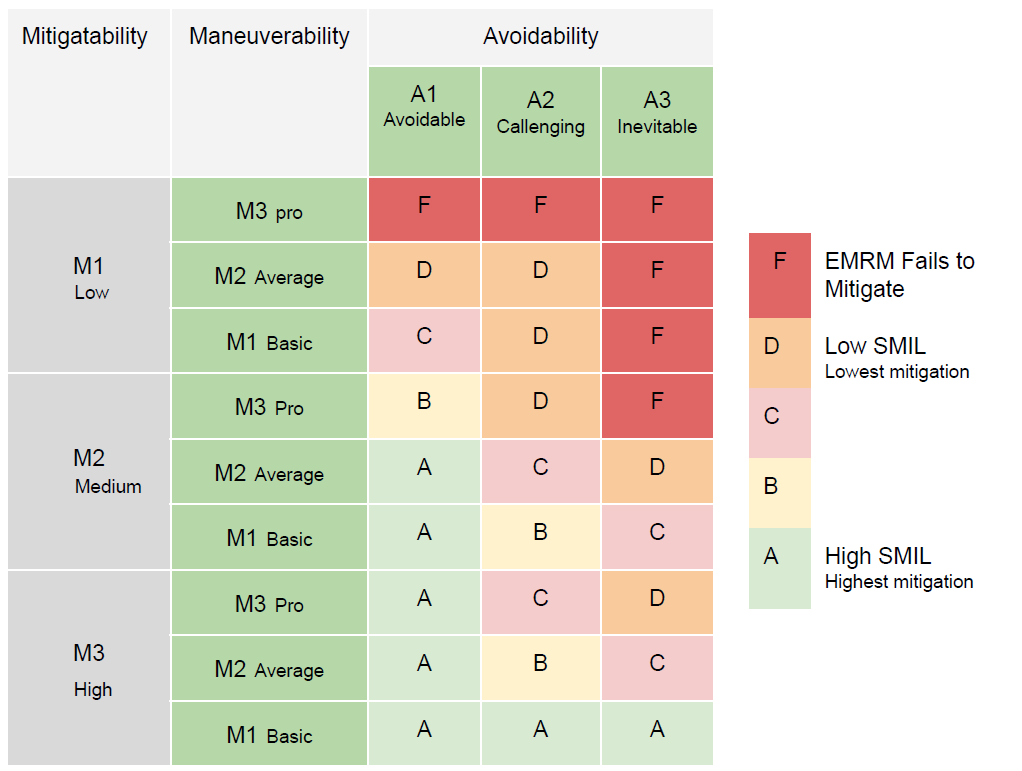}
	\label{table:SMIL_Table}
 \vspace{-0mm}
\end{table}

\subsection{Loss Evaluations}\label{subsec:Loss Evaluation Procedure}
The loss evaluation procedure, as the final part of safety verification for an EMRM system, involves assessing driving skills required to mitigate a loss, the probability of successful mitigation, and the impact of the loss if it occurs based on the definitions in Section ~\ref{subsec:Loss Evaluation Parameters}. This process evaluates how effectively hazards can be managed based on the the effectiveness of evasive maneuvering, ultimately determining the potential consequences and the overall risk level associated EMRM systems being used in critical hazardous situations. Out of $12$ hazardous scenarios in Table~\ref{table:List of hazardous scenarios}, $7$ are considered as QM where the chance of severe loss is noticeable. Further loss evaluations is necessary to verify if a skilled evasive maneuver might be able to reduce the loss probability or impact. \textcolor{black}{Table~\ref{table:Loss Evaluation} shows the loss reduction evaluation and a mitigation Synopsis for every critical hazard in Table~\ref{table:List of hazardous scenarios}}. It is important to note, since the analysis are only for the catastrophic and critical hazards, the ODD is for high-speed driving situations and critical road and complex scenarios, similar to Fig.~\ref{fig:realscenario}.

\subsubsection{Safety Goals}\label{subsec:Loss Evaluation Procedure}
\textcolor{black}{Table~\ref{table:SMIL_Table} shows the loss evaluations for ASIL QM hazardous in Table~\ref{table:List of hazardous scenarios}. This table includes the scenarios of Severity Mitigation Integrity Level (SMIL) mapping for EMRM systems.} An EMRM system will be capable of the loss mitigation in higher SMIL scenarios and the chance of loss mitigation will drop for lower level SMIL scenarios. There are certain scenarios that EMRM will most likely fail and it not justified to design an EMRM system for these kind of scenarios. The safety goal for an EMRM system is to activate aggressive maneuvers only when there is a very high probability of successful loss mitigation, ensuring that the maneuver is stable and the potential loss is avoidable. SMIL level B and D are considered as acceptable scenarios for using EMRM,\textcolor{black}{where H5 and H6 are not suitable scenarios to use EMRM.}

\begin{table*}[ht!]
\centering
\caption{Loss reduction evaluations and mitigation synopsis for the critical and catastrophic hazards in Table~\ref{table:List of hazardous scenarios}.}
\label{table:Loss Evaluation}
\begin{tabular}{|c|c|>{\centering\arraybackslash}p{0.61\linewidth}|c|c|c|c|}
\hline
\textbf{$\mathcal{L}$ID} & \textbf{HID} & \textbf{Evasive MRM Scenario} & \textbf{Manu.} & \textbf{Avoi.} & \textbf{Mit.} & \textbf{SMIL} \\ \hline
$\mathcal{L}$1 & H1 &  EMRM system performs aggressive maneuvers to reduce accident impact and loss severity on the VRU due to inability to detect the VRU properly. & M$_3$ & A$_2$ & $\mathcal{M}_1$ & D \\ \hline
$\mathcal{L}$2 & H2 & EMRM system performs hard braking and turning direction to avoid an object due to False-Positive detection. & M$_3$ & A$_2$ & $\mathcal{M}_2$ & B \\ \hline
$\mathcal{L}$3 & H4 & EMRM system performs aggressive maneuvers immediately to reduce accident impact and loss severity for the ego and other vehicles. & M$_3$ & A$_3$ & $\mathcal{M}_3$ & D \\ \hline
$\mathcal{L}$4 & H5 & EMRM perform non-aggressive maneuvers to reduce accident impact due to wrong loss predictions. & M$_3$ & A$_3$ & $\mathcal{M}_2$ & F\\ \hline
$\mathcal{L}$5 & H6 & EMRM system performs hard braking and/or turning direction to avoid an object due to false severity identification. & M$_2$ & A$_1$ & $\mathcal{M}_1$ & F \\ \hline
$\mathcal{L}$6 & H7 & EMRM perform less evasive maneuvers to reduce accident impact due to system limits. & M$_3$ & A$_2$ & $\mathcal{M}_2$ & D\\ \hline
$\mathcal{L}$7 & H9 & EMRM perform maneuvers with delay to reduce accident impact due to system limits. & M$_3$ & A$_2$ & $\mathcal{M}_1$ & D\\ \hline
\end{tabular}
\end{table*}

\textcolor{black}{\subsection{Impact of System and Environmental Factors}
\subsubsection{Sensors and Perception}
\noindent Sensors data and perception modules significantly impact the hazard detection. High-resolution LIDAR, RADAR, and cameras improve perception and high-speed and high precision perception modules provide timely and accurate information, thereby lowering the probability ($P'$) of the hazard by enabling early detection and preemptive maneuvers ~\cite{levinson2010robust}. For example, sensors and cameras data consumed by a perception algorithms for detecting a pedestrian in the vehicle's path allow for an emergency swerve or controlled slide, reducing collision likelihood ~\cite{katrakazas2015real}. Obviously, Any inaccuracy in these systems will drastically increase the risk level where redundant design is highly recommended in sensor and perception architectures. 
\subsubsection{Predictive and Decision Making}
\noindent Predictive models, planning algorithms and controllers effect the system's controllablity to avoid a hazardous situation. However, the vehicle’s mechanical integrity, such as the state of tires, brakes and batteries, affects the maneuver’s success and thus the risk level. Quick and accurate responses to perception alerts, potentially enhanced by predictive models and decision making systems, can further reduce risk~\cite{Wang2022Risk}. Any inaccuracy in predictive and decision making models will drastically increase the risk level. 
\subsubsection{Environmental and Traffic}
\noindent Environmental conditions, such as weather and road surface, affect maneuverability and subsequently the exposure ($E'$) and severity ($S'$) of hazards. Adverse conditions like rain, snow, or fog increase these factors, while favorable conditions reduce them, enhancing the effectiveness of EMRMs. Poor environmental conditions often lead to higher severity and exposure, as the likelihood of losing control during maneuvers rises ~\cite{qiu2008effects}. Also,  Dynamic factors such as vehicle speed, traffic density, and the vehicle's condition also play a crucial role~\cite{hsiao2018preventing}. Higher speeds and denser traffic elevate severity and exposure, necessitating swift and precise EMRMs.}

\section{Conclusion}\label{sec:conclusion}
This paper introduced a novel safety verification methods for the EMRM feature of AVs. EMRM systems are designed to handle  high risk, low probability scenarios. To verify the safety of an EMRM system, it is essential to analyze potential losses alongside hazard risks. The method's formulation, along with a detailed case study on advanced EMRM, highlights its safety goals. By leveraging maneuvering skills from professional racing and stunt driving, combined with machine learning techniques and formal safe control methods, the safety verification approach effectively evaluates safety goals for using evasive maneuvers in complex hazards, providing robust collision avoidance in critical situations. The proposed framework can serve as a baseline for further safety analysis and validation of EMRM systems. Our future work will involve applying these methods to real-world scenarios and refining safety metrics. \textcolor{black}{This work primarily focused on external entities and hazards.
However, the safety of the occupants themselves are a crucial part of the hazard analysis. As a future research direction, we enhance this framework by considering both internal and external hazards to provide a comprehensive risks management strategy. Future work also includes adoption of IEEE P3332 and IEEE P2817 for system-level hazard analysis.}

\bibliographystyle{IEEEtran}
\bibliography{IEEEabrv,ArabRef}
\vfill
\end{document}